\pdfoutput=1
\documentclass[11pt]{article}

\usepackage[final]{acl}

\usepackage{times}
\usepackage{latexsym}

\usepackage[T1]{fontenc}
\usepackage[utf8]{inputenc}

\usepackage{microtype}

\usepackage{inconsolata}

\usepackage{graphicx}

\usepackage{amsmath}
\usepackage{amssymb}
\usepackage{booktabs}
\usepackage{multirow}
\usepackage{array}
\usepackage{makecell}
\newcommand{\mymodel}{D\&R}

\title{Debate, Reflect, and Distill: Multi-Agent Feedback with Tree-Structured Preference Optimization for Efficient Language Model Enhancement}

\author{
 \textbf{Xiaofeng Zhou\textsuperscript{1}\thanks{This work was done during an internship at SMU.}},
 \textbf{Heyan Huang\textsuperscript{1}},
 \textbf{Lizi Liao\textsuperscript{2}}
\\
\\
 \textsuperscript{1}Beijing Institute of Technology,
 \textsuperscript{2}Singapore Management University
\\
 \texttt{\{zhouxiaofeng, hhy63\}@bit.edu.cn, lzliao@smu.edu.sg}
}

\begin{document}
\maketitle
\begin{abstract}
Large Language Models (LLMs) continue to set new standards in knowledge-intensive and complex reasoning tasks, yet their high computational demands limit widespread adoption. While distilling large models into smaller ones offers a sustainable solution, current techniques—such as static knowledge distillation, resource-intensive reinforcement learning from human feedback, or limited self-reflection—struggle to yield substantial and lasting performance gains. In this paper, we present a novel \textbf{D}ebate and \textbf{R}eflect (\textbf{D\&R}) framework that orchestrates multi-turn debates between smaller models and stronger teacher models, eliciting actionable feedback (\textit{e.g.}, error analysis, corrective strategies) to guide student models. Further, we introduce Tree-structured Direct Preference Optimization (T-DPO) to efficiently leverage these debate logs, organizing interactions into a hierarchical format for effective training. Empirical evaluations across diverse NLP benchmarks demonstrate that our approach significantly improves smaller-model accuracy, robustness, and generalization, outperforming conventional baselines by a large margin.
\end{abstract}

\section{Introduction}
LLMs have emerged as powerful tools for tackling complex reasoning tasks \citep{zelikman_quiet-star_2024, qwen_qwen25_2025, he_planning_2024}. Their ability to process and generate coherent, context-aware outputs has set a new standard in artificial intelligence (AI). However, the computational demands of training and deploying these models make them inaccessible to many users, highlighting the need for more efficient alternatives. Developing smaller models capable of achieving similar knowledge and reasoning performance is an essential step toward making advanced AI technologies more accessible and sustainable.

\begin{figure}
    \centering
    \includegraphics[width=0.99\columnwidth]{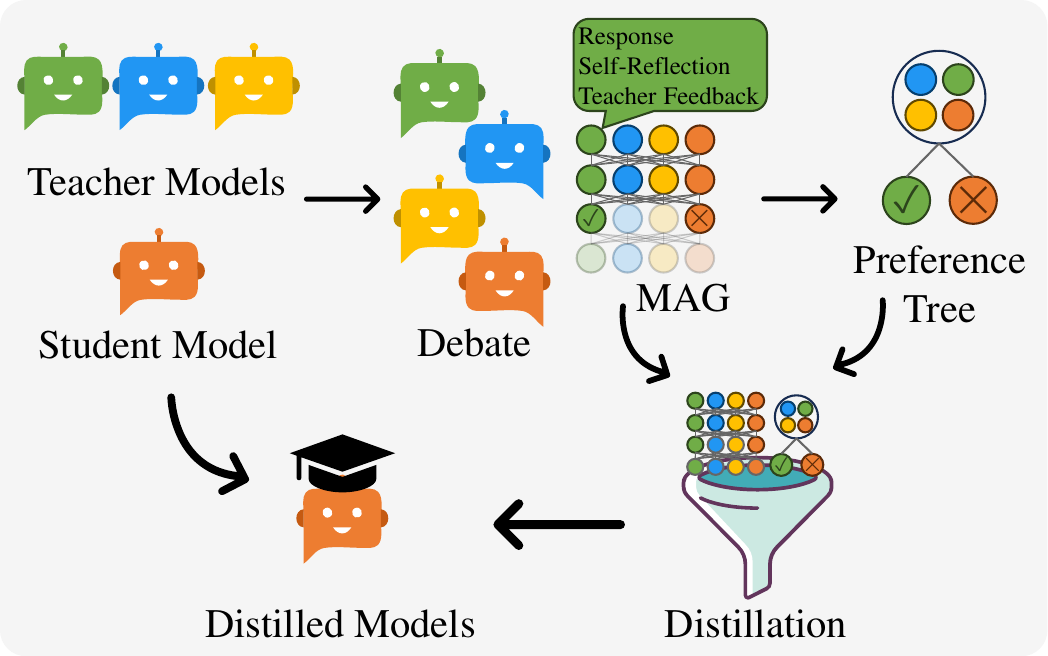}
    \caption{Overview of \mymodel. The student model debates with teacher models, receiving 
self-reflection and teacher feedback. A Multi-Agent Interaction Graph (MAG) 
records debate content, from which preference trees are extracted. 
All information is then distilled into the student, enhancing its 
knowledge and reasoning.}
\vspace{-0.4cm}
    \label{fig:procedure}
\end{figure}

Existing strategies for enhancing smaller models generally fall into three categories. First, \textbf{knowledge distillation} transfers knowledge from a larger teacher to a smaller student model \citep{hinton_distilling_2015, shridhar_distilling_2023}. While effective for reducing model size, traditional distillation procedures are typically one-off processes that lack targeted guidance from the teacher. That is, the teacher’s rich insights into how the student can improve are not fully leveraged, and the student often receives only generic training signals. Second, \textbf{reinforcement learning from human feedback} (RLHF) aligns model behavior with human preferences and can substantially boost reasoning \citep{stiennon_learning_2020, liang_bit_numeval_2024}. However, it requires extensive human intervention, raising concerns about scalability and cost. Third, \textbf{multi-agent systems and self-reflection} frameworks enable models to exchange critiques or engage in introspective refinement during inference \citep{liang_encouraging_2024, madaan_self-refine_2023, wang_qrmem_2024}. Although these methods show promise in improving model reasoning, they often apply feedback only transiently—without permanently updating model weights—or use simplistic fine-tuning schemes that do not fully exploit iterative feedback.

In this paper, we introduce a novel \textbf{D}ebate and \textbf{R}eflect (\textbf{D\&R}) framework that combines the strengths of multi-agent debate and actionable teacher feedback to robustly enhance smaller models. Concretely, a student model engages in multi-turn debates with one or more stronger teacher models, collecting detailed feedback in the form of error analyses, corrective strategies, and best practices. Crucially, the feedback is not limited to a single inference; rather, we incorporate both \textbf{self-reflection} and \textbf{teacher feedback} mechanisms that help the student to internalize these insights for lasting, parameter-level improvements \citep{yuan_self-rewarding_2024, ying_llms-as-instructors_2024, lan_training_2024}. To maximize the efficiency of this training signal, we further propose Tree-structured Direct Preference Optimization (T-DPO) to transform debate logs into hierarchical preference trees \citep{yuan_advancing_2025}. By systematically organizing intermediate debate turns and teacher guidance, T-DPO enables the student to learn not just what the best response is, but why certain answers are preferred. Through extensive empirical evaluations, we show that D\&R with T-DPO significantly outperforms conventional methods in accuracy, robustness, and generalization, offering a promising route for achieving high performance in smaller, more accessible language models.

In summary, our contributions are threefold:
\begin{itemize}

\item  We introduce a Debate and Reflect paradigm that combines the strengths of multi-agent interaction and targeted teacher feedback to continuously improve smaller models.
\item We propose a Tree-structured Direct Preference Optimization (T-DPO) to better leverages debate data by structuring it into preference trees for efficient and robust training.
\item Extensive experiments across diverse NLP tasks, demonstrating that our method significantly outperforms standard knowledge distillation and supervised fine-tuning.

\end{itemize}

\begin{figure*}
    \centering
    \includegraphics[width=0.99\linewidth]{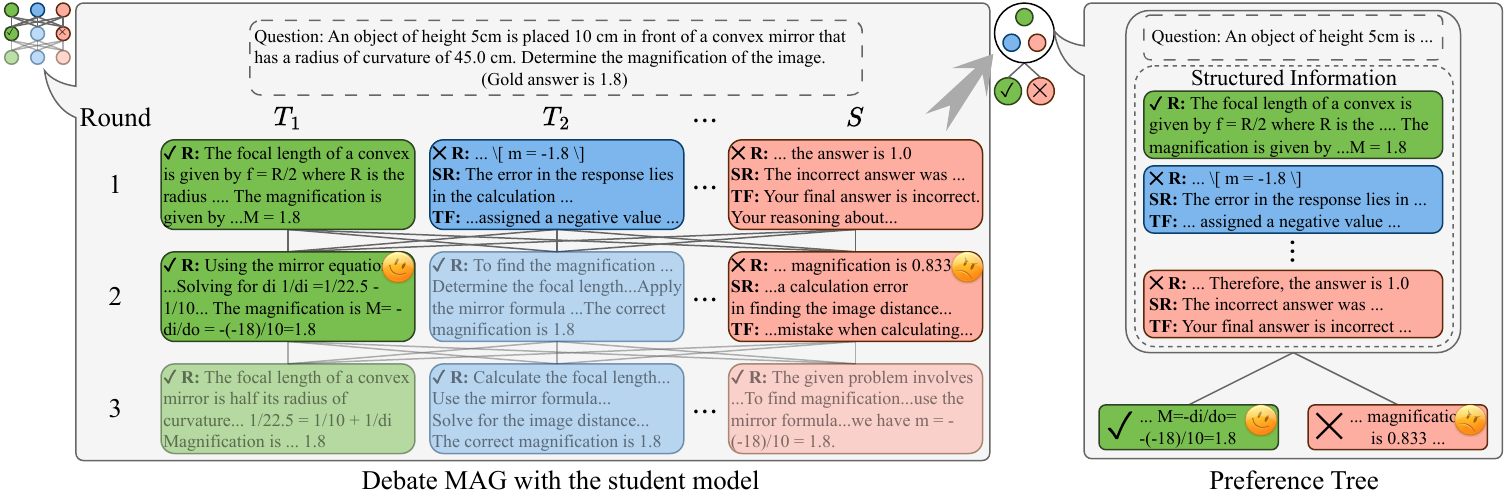}
    \vspace{-0.2cm}
    \caption{A debate example of our proposed \mymodel. The left part demonstrates the MAG of a debate involving the teacher models $T_{1}, T_{2}, \cdots$ and the student model $S$. Each node corresponds to an agent in a round, including responses (R), self-reflection (SR), and teacher feedback (TF). The highlighted nodes are extracted to construct a preference tree for T-DPO. In its root node, there is the structured information representing the previous debate.}
    \label{fig:overview}
    \vspace{-0.5cm}
\end{figure*}

\section{Related Work}
\subsection{Knowledge Distillation}
Knowledge distillation \citep{hinton_distilling_2015} is a widely used technique for transferring knowledge from large and high-performing teacher models to smaller student models, enhancing performance while reducing inference costs. Given the varying parameter sizes, capabilities, and computational requirements of LLMs \citep{touvron_llama_2023-1, jiang_mistral_2023, li_fundamental_2024, zhang_analyzing_2024, wang_boosting_2023, ren_tcm-sd_2022, zhou_rs-bert_2024, bai_citrus_2024}, distilling knowledge from powerful teacher models is a common strategy for optimizing the trade-off between performance and efficiency. Typically, knowledge distillation involves querying teacher models with a set of seed data to extract relevant knowledge, which is then used to train student models through various distillation algorithms \citep{xu_wizardlm_2024, gu_minillm_2024}. In the context of reasoning tasks, Chain of Thought (CoT) reasoning serves as a crucial learning signal, allowing student models to learn structured reasoning steps from their teachers \citep{mukherjee_orca_2023}. Beyond CoT, feedback has also been employed to provide process supervision, further refining student models’ reasoning capabilities \citep{luo_wizardmath_2023}. However, conventional knowledge distillation methods are typically static, meaning they lack iterative interactions between teacher and student models. To address this, our work utilizes CoT and feedback from debate-based interactions, providing more targeted and dynamic guidance for student model improvement.

\subsection{Multi-Agent Debate}
Multi-agent debate refines LLM reasoning and decision-making based on LLMs' powerful conversational ability \citep{chu_towards_2024, liang_survey_2024, yang_building_2024} and agents' collaborative ability \cite{li_camel_2023, tu_spagent_2024}, by exploring diverse reasoning paths and cross-verifying claims, thereby reducing hallucinations and errors \citep{du_improving_2024}. It also serves as an evaluation mechanism for tasks like question answering and summarization \citep{chan_chateval_2024}. Debate formats vary—from collaborative \citep{xiong_examining_2023} to adversarial \citep{cohen_lm_2023}—with different communication protocols \citep{chan_chateval_2024}. Recent studies show that such frameworks can even drive self-improvement without human supervision \citep{fu_improving_2023}. For instance, \textsc{MAGDi} \citep{chen_magdi_2024} uses multi-teacher debates to build interaction graphs that yield richer learning signals than single-teacher setups. In our work, we leverage multi-agent debates with stronger models to generate richer reasoning and actionable feedback, integrating these insights into our Tree-structured Direct Preference Optimization pipeline.

\subsection{DPO and Reasoning}
Direct Preference Optimization (DPO) \citep{rafailov_direct_2023} offers a stable and computationally efficient way to align LLMs with specific objectives, making it valuable for refining both behavior and reasoning quality. Several extensions illustrate how DPO can be harnessed to improve reasoning. For instance, RPO \citep{pang_iterative_2024} iteratively applies DPO alongside negative log-likelihood losses to fine-tune models on complex tasks. CPO \citep{zhang_chain_2024} and Eurus \citep{yuan_advancing_2025} take a tree-structured approach, building hierarchical reasoning paths to guide DPO-based alignment. Further refinements like Step-DPO \citep{lai_step-dpo_2024} and Step-Controlled DPO \citep{lu_step-controlled_2024} apply alignment at each individual reasoning step, ensuring that the model learns correct transitions between steps. In addition, M-DPO \citep{xiong_building_2025} shows how DPO can be extended to multi-turn dialogues, incorporating tool usage for solving math problems. In our work, we extend DPO with a Tree-structured approach (T-DPO) that turns debate logs into hierarchical preference trees, enabling the student model to learn from both final answers and the intermediate reasoning in multi-turn debates.

\section{Debate \& Reflect Method}
To improve small models using \textbf{D}ebate \& \textbf{R}eflect (\textbf{\mymodel}), we construct debates where the small student model engages in discussions with their teacher models, receiving feedback from them (Section \ref{method_debate}). From debates, we construct preference trees in Multi-Agent Interaction Graphs (MAGs) including debate context for T-DPO (Section \ref{method_tree}). Finally, we distill these structured insights into the student model with T-DPO, enabling more effective learning (Section \ref{method_distillation}).

\subsection{Small Models Engage in Debates} \label{method_debate}
To enable iterative interaction and targeted feedback, we design a debate framework where a student model engages in discussions with stronger teacher models. Figure \ref{fig:overview} shows a debate example. We use training sets as our data sources to obtain gold answers. During the debates, multiple agents engage in several rounds of discussion on the same problem, exchanging their insights, and ultimately achieving a consensus answer. In each debate round, the teacher models $\mathcal{T}=\{T_{i}\}^{n}_{i=1}$ and the student model $S$ analyze the structured information from the previous debate round. This information includes prior responses, along with the corresponding self-reflection and feedback. Based on this analysis, each agent generates a new response that includes both a reasoning process and a final answer. To evaluate performance, we compare each agent’s response against the gold answer. Agents who get an incorrect answer (such as the student model $S$) are prompted to self-reflect, while agents with correct answers (such as a teacher model $T_{i}$) provide constructive feedback to incorrect agents. Teacher models actively highlight student errors and suggest corrections, while the student model attempts to recognize and address its own mistakes. Then the newly generated responses, including reasoning processes, self-reflections, and feedback, are passed to the next round as updated structured information. In the next round, the models also analyze the strengths and weaknesses of their previous responses again, identifying potential errors and incorporating corrective feedback to generate next-round responses. This iterative process enables the student model to progressively refine its reasoning and responses. Debates continue until all agents reach a consensus or the maximum number of rounds is reached. Following \citet{chen_magdi_2024}, we construct MAGs for debates, which record debate processes as structured graphs for  distillation.

\subsection{Construct Preference Trees for T-DPO} \label{method_tree}
To effectively train models on reasoning trajectories, we construct preference trees as structured representations of debate interactions. Inspired by previous studies \citep{yuan_advancing_2025, zhang_chain_2024}, we propose Tree-structured DPO (T-DPO) for debate distillation, which constructs preference trees from graphs for preference optimization. 

The right hand side of Figure \ref{fig:overview} shows an example of our preference tree. Preference trees capture small, specific reasoning cases from debates. In the debates mentioned above, agents receive structured information from the previous round and generate new responses. Each agent is given the same input prompt in the same round, containing the problem and the previous structured information. However, due to varying capabilities, agents produce different responses; some are correct, and others are incorrect. Therefore, we choose an input prompt as the root node of a preference tree. A correct response and an incorrect response derived from this prompt become child nodes. The root node stores not only the previous round's responses but also self-reflection and feedback on the incorrect answers. This allows agents to integrate insights from past correct responses while addressing mistakes through refinement. Responses that evolve into correct answers (chosen nodes) illustrate how effective reasoning develops over multiple debate rounds. Conversely, responses that remain incorrect (rejected nodes) highlight lessons that should not be reinforced in the distillation process. Through this hierarchical structure, preference trees systematically map out reasoning trajectories, helping models internalize both successful strategies and critical learning points from past errors.

\subsection{Distillation} \label{method_distillation}
Our distillation pipeline consists of two key stages: Supervised Fine-Tuning (SFT) and Tree-structured DPO (T-DPO). First, we extract problems and their corresponding correct responses from MAGs to construct the SFT dataset. Each data instance in this dataset includes an instruction $I$, a problem $x_{i}$, and its correct answer $y_{i}$, but excludes the structured information from the previous debate round. The student model is optimized with maximum likelihood estimation:
\begin{equation}
    \nonumber
    \mathcal{L}_{SFT}=-\mathbb{E}_{(I, x, y)\sim \mathcal{D}_{SFT}} \log \pi_{\theta} (y|I, x),
\end{equation}
where $\mathcal{D}_{SFT}$ is the SFT training dataset and the student model $\pi_{\theta}$ is initialized as the original small model before distillation.

In the second stage, we apply T-DPO training to refine the student model’s reasoning. The root node of each preference tree serves as the input prompt, incorporating the instruction $I$, the problem $x$, and the previous structured information $SI$. The correct and incorrect child nodes act as the chosen and rejected responses, $y_{w}$ and $y_{l}$, respectively. The student model is optimized with the T-DPO objective:
\begin{align}
    \nonumber
    \mathcal{L}_{T-DPO}=&-\mathbb{E}_{(I, x, SI, y_w, y_l)\sim \mathcal{D}_{T-DPO}}\\
    \nonumber
    &\left[\log \sigma \left(\beta \log \frac{\pi_{\theta}(y_w|I, x, SI)}{\pi_{ref}(y_w|I, x, SI)} \right.\right.\\
    \nonumber
    &\left.\left. - \beta \log \frac{\pi_{\theta}(y_l|I, x, SI)}{\pi_{ref}(y_l|I, x, SI)}\right)\right],
\end{align}
where $\mathcal{D}_{T-DPO}$ is all preference trees in the T-DPO training dataset, $\sigma$ is the logistic function, $\pi_{\theta}$ is the student model policy, $\pi_{ref}$ is the base reference policy, and $\beta$ is the hyperparameter controlling the deviation from the base reference policy. Both $\pi_{\theta}$ and $\pi_{ref}$ are initialized with the above SFT-trained student model.

After undergoing SFT and T-DPO, the student model acquires the ability to construct correct reasoning paths while effectively avoiding past errors. During inference, the model processes problems step by step, systematically generating thoughts and analyses, ultimately leading to final answers.

\section{Experiments}

\subsection{Experimental Settings}
This section outlines the major settings. More setting details in Appendix \ref{app_experimental_setting_details}.

\paragraph{Datasets.}
The MMLU Pro benchmark \citep{wang_mmlu-pro_2024} is a multi-discipline language understanding and reasoning benchmark, extending the MMLU benchmark \citep{hendrycks_measuring_2021} with more challenging problems. The dataset is partially derived from MMLU, with additional samples sourced from the Internet and other datasets. It not only evaluates the models' knowledge but also emphasizes reasoning capabilities. Notably, it expands the number of answer choices to 10. For our experiments, we choose three categories of MMLU Pro: computer science, physics, and biology. These categories vary in data size and complexity, making them ideal for evaluating both knowledge retention and logical reasoning. Since MMLU Pro does not provide a dedicated training set, we split the original test set into separate training and test sets. Statistical information of datasets is shown in Table \ref{tbl:data_stat}.

The MATH dataset \citep{hendrycks_measuring_2021-1} is a benchmark designed to assess step-by-step mathematical reasoning in LLMs. It requires LLMs to think about the problems step by step and generate a final answer, such as a mathematical expression. In our experiments, we adopt the training data in \citet{chen_magdi_2024} but reconstruct the debates and MAGs with our teacher and student models.

\paragraph{Baselines.}
To evaluate our method, we compare it against various distillation baselines. (1) \textit{No Distillation}: the original Mistral-7B-Instruct model without any distillation. For MMLU Pro, it is tested in a zero-shot setting. For MATH, we provide only a few simple demonstrations from \cite{hendrycks_measuring_2021-1} to illustrate the expected output format. (2) \textit{Single-Teacher Distillation}: The student model is distilled using SFT with knowledge from only one teacher model at a time. (3) \textit{Multi-Teacher Distillation}: \textsc{MAGDi} \citep{chen_magdi_2024} uses both correct and incorrect answers of multiple teacher models for distillation. But it does not involve the student interacting with teachers and lacks reflection or feedback mechanisms. Following the official code of \textsc{MAGDi}, we reimplement it with our debate data. For all baseline models and \mymodel, We select GPT-4o\footnote{\url{https://openai.com/index/hello-gpt-4o}}, Claude 3.5\footnote{\url{https://www.anthropic.com/claude/sonnet}}, and Gemini 1.5 Pro\footnote{\url{https://ai.google.dev/gemini-api/docs/models/gemini}} as our teacher models, while the student model is Mistral-7B-Instruct \citep{jiang_mistral_2023}. 

\paragraph{Evaluation Metrics.}
For MMLU Pro, we use accuracy as the evaluation metric. For MATH, we follow \citet{hendrycks_measuring_2021-1} to use exact match after normalizing answers to get accuracy.

\paragraph{Implement Details.}
To ensure high-quality distillation while controlling computational costs, we impose the following settings.\footnote{We release our code in \url{https://github.com/zhouxiaofengshelf/D-R}.} When constructing debates, we limit debates to a maximum of 4 rounds to balance cost-efficiency and data richness. When collecting root node data for preference trees, we control the length of structured information since it is impractical to train the student model with too long sequences. We aim to include as much structured information in root nodes as possible and place the remainder into a new preference tree. For the first round, the preference trees have no structured information from the previous round. During distillation, we implement our distillation with LoRA \citep{hu_lora_2022}. The prompt templates used in our experiments are shown in Appendix \ref{app_temp}.

\begin{table*}
  \centering
  \small
  \begin{tabular}{l*{2}{c}ccc}
      \toprule
      \multirow{2}{*}{Models} & \multicolumn{3}{c}{MMLU Pro} & \multirow{2}{*}{MATH} & \multirow{2}{*}{Average} \\
      & Computer Science (CS) & Physics & Biology & \\
      \midrule
      \multicolumn{6}{l}{\textbf{\textit{No Distillation}}} \\
      \midrule
      Mistral-7B-Instruct & 20.49 & 18.46 & 48.95 & 8.02 & 23.98 \\
      \midrule
      \multicolumn{6}{l}{\textbf{\textit{Single-Teacher Distillation}}} \\
      \midrule
      Single Teacher$_{GPT-4o}$ & 27.80 & 28.63 & 62.34 & 16.92 & 33.92 \\
      Single Teacher$_{Claude\ 3.5}$ & 30.73 & 30.77 & 62.76 & 16.56 & 35.21 \\
      Single Teacher$_{Gemini\ 1.5\ Pro}$ & 25.85 & 24.92 & 60.67 & 15.86 & 31.83 \\
      \midrule
      \multicolumn{6}{l}{\textbf{\textit{Multi-Teaher Distillation}}} \\
      \midrule
      \textsc{MAGDi} & 22.93 & 20.00 & 54.39 & 11.26 & 27.15 \\
      \mymodel$_{SFT}$ & 32.68 & 29.23 & 64.44 & \textbf{17.64} & 36.00 \\
      \mymodel & \textbf{33.17} & \textbf{34.77} & \textbf{67.36} & 17.32 & \textbf{38.16} \\
      \bottomrule
  \end{tabular}
  \caption{\label{tbl:main_results}
    The main results of our experiments on three categories of MMLU Pro and MATH. We compare our \mymodel~with three types of distillation baselines and SFT-only \mymodel$_{SFT}$. \textsc{MAGDi} are reimplemented with its official code and our debate data. \textbf{Bold} figures denote the best results.
  }
  \vspace{-0.3cm}
\end{table*}

\subsection{Main Results.}
We show our experiment results in Table \ref{tbl:main_results}. The student model's knowledge and reasoning capability across three disciplines, computer science, physics, and biology, are evaluated by MMLU Pro. The complex mathematic reasoning capability is evaluated by MATH.

\paragraph{\mymodel~significantly enhances the student model through distillation.}
Before distillation, the original Mistral-7B-Instruct demonstrates moderate performance. On MMLU Pro, it surpasses the random guessing performance (10.00) by a meaningful margin, highlighting its basic knowledge and reasoning capability. However, on MATH, it achieves only 8.02, indicating that it can handle complex reasoning for only a small subset of MATH problems. All distillation methods lead to notable improvements over the original Mistral-7B-Instruct. \mymodel~gets an average improvement of 14.18 (up to 18.41 in the biology category of MMLU Pro). This strong improvement underscores \mymodel's ability to capture the valuable insights of debates and recognize the preference between the correct and incorrect answers.

\paragraph{\mymodel~outperforms Single-Teacher Distillation.}
As baseline models, single-teacher distillation results vary across different teacher models. Among them, Single Teacher$_{Claude\ 3.5}$ achieves the best average results of single-teacher distillation, while Single Teacher$_{Gemini\ 1.5\ Pro}$ lags behind, with an average performance gap of 3.38. Despite these variations, all single-teacher distillation methods lead to meaningful improvements, proving that knowledge and reasoning can be effectively transferred from teacher models.

Compared to the single-teacher distillation, \mymodel~achieves better distillation results. On the average performance, \mymodel~improves at least 2.95 (from 35.21 of Single Teacher$_{Claude\ 3.5}$ to 38.16 of \mymodel). The key advantages of \mymodel~include multiple teacher models for richer insights, debate mechanisms for iterative interactions, and T-DPO for structured reasoning preference learning. These factors collectively enable more effective knowledge and reasoning acquisition. The debate process enhances training data quality, focusing on the student model’s weaknesses, while T-DPO refines learning by emphasizing the contrast between correct and incorrect responses.
\vspace{-0.2cm}

\paragraph{\mymodel~achieves the best performance in Multi-Teacher Distillation.}
\mymodel$_{SFT}$ already achieves strong performance, surpassing most baseline models. T-DPO further enhances it, leading to an additional 2.16 improvement in the average performance. On the MATH dataset, the result of \mymodel~is slightly lower than the one of \mymodel$_{SFT}$ but still higher than all other baseline models. We will conduct further exploration in Section \ref{exp_stu_model}. Compared to \textsc{MAGDi}, our \mymodel~demonstrates a substantial average gain of 11.01 (up to 14.77 in the physics category of MMLU Pro). While both \mymodel~and \textsc{MAGDi} utilize multiple teachers and debates, \mymodel~generates richer debate content. Involving the student model in the debates can lead to higher-quality training data. The teacher models are aware of errors in the student model and can actively provide explicit feedback. The student model learns not only from teacher responses but also from the feedback and their own self-reflection to know where and why it makes errors. This process makes \mymodel's distillation approach superior to passively learning from teacher responses without explicit error analysis, as done in \textsc{MAGDi}. Moreover, T-DPO provides a more structured and effective optimization framework than the margin-based objective \citep{cortes_support-vector_1995} and GCN \citep{kipf_semi-supervised_2017} of \textsc{MAGDi}. \textsc{MAGDi} underperforms compared to the single-teacher distillation, suggesting its joint learning of three training objectives may not model the debate data effectively.

\begin{figure}
    \centering
    \includegraphics[width=0.9\columnwidth]{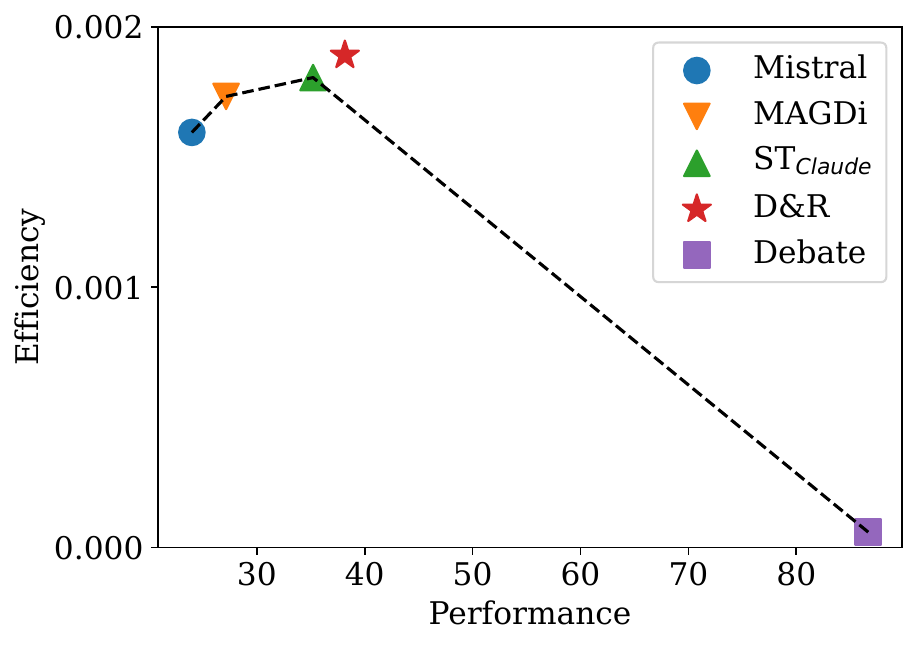}
    \vspace{-0.2cm}
    \caption{Trade-off between performance and efficiency. Inspired by \citet{chen_magdi_2024}, we define efficiency as $1/avg(tokens)$. \mymodel~improves both performance and efficiency compared to previous distillation methods, surpassing their Pareto frontier. We choose Single Teacher$_{Claude\ 3.5}$ (ST$_{Claude}$) to represent the Single-Teacher Distillation.}
    \label{fig:perf_effi}
    \vspace{-0.4cm}
\end{figure}

\subsection{Detailed Analysis}

\subsubsection{\mymodel~Improves Inference Efficiency} \label{exp_efficiency}
\mymodel~effectively balance efficiency and performance. Although LLMs and debate can provide strong knowledge and reasoning, they come at a high inference cost. Distillation improves efficiency, making reasoning models more accessible and sustainable. Inspired by \citet{chen_magdi_2024}, we measure inference efficiency by calculating token cost per problem across three categories in MMLU Pro (excluding MATH due to API cost). We report our detailed token cost in Appendix \ref{app_token_cost}. Figure \ref{fig:perf_effi} illustrates the trade-off between performance and efficiency. Debate incurs the highest token cost, making it the least efficient. After distillation, \textsc{MAGDi} and Single Teacher$_{Claude\ 3.5}$ can generate better reasoning paths, improving their performance with high efficiency. \mymodel~surpasses the boundaries of previous distillation methods, not only improving performance but also effectively enhancing efficiency. It becomes a new optimal balance of performance and inference efficiency.

\begin{table}
  \centering
  \small
  \begin{tabular}{llc}
      \toprule
      Models & \makecell[l]{Additional\\Correction} & Accuracy \\
      \midrule
      \mymodel & None & 34.77 \\
      \midrule
      \mymodel & \multirow{4}{*}{Self-Reflect} & \textbf{36.92} \\
      \hspace{2ex}without SR data && 33.85 (-3.07) \\
      \hspace{2ex}without TF data && 32.31 (-4.61) \\
      \hspace{2ex}without SR \& TF data && 31.38 (-5.54) \\
      \midrule
      \mymodel & \multirow{4}{*}{\makecell[l]{Teacher$^*$\\Feedback}} & \textbf{37.54} \\
      \hspace{2ex}without SR data && 32.62 (-4.92) \\
      \hspace{2ex}without TF data && 32.00 (-5.54) \\
      \hspace{2ex}without SR \& TF data && 31.08 (-6.46) \\
      \bottomrule
      \multicolumn{3}{p{7cm}}{\scriptsize * As during inference, we do not allow teacher models to involve, we let the student model role-play a teacher.
    }
  \end{tabular}
  \vspace{-0.2cm}
  \caption{\label{tbl:sr_tf_boost}
  Effect of self-reflection (SR) and teacher feedback (TF) in D\&R. We also show D\&R with additional self-reflection and teacher feedback during inference.
  }
  \vspace{-0.4cm}
\end{table}

\subsubsection{\mymodel~Can Correct Itself}
\mymodel~not only improves reasoning but also learns to identify and correct its own mistakes. In our debates, agents are explicitly required to generate self-reflection and feedback to identify and correct errors. During distillation, the student model learns this correction process. To evaluate its error correction ability, we conduct an additional correction step during inference. In correspondence with the self-reflection and feedback in the training debates, we design two additional correction methods. The first one is self-reflection. After generating an initial response, the student model self-reflects on its own response and updates its answer. The second one is teacher feedback simulation. The student model roleplays a teacher model, critiques its own response as if reviewing another agent’s answer, and then corrects errors (both the teacher and student models are actually the same model). Table \ref{tbl:sr_tf_boost} shows how self-reflection and teacher feedback in training debates improve the student model's performance. With these additional correction methods, \mymodel~can leverage its learned error correction ability to refine its reasoning. However, without the self-reflection and feedback data from the training debates for distillation, the student model fails to effectively self-correct or simulate teacher feedback during inference. This highlights that beyond reasoning, self-reflection and teacher feedback in training debates are essential for transferring critical correction abilities to the student model, enhancing its performance through additional correction at inference.

\begin{figure}
    \centering
    \includegraphics[width=0.9\columnwidth]{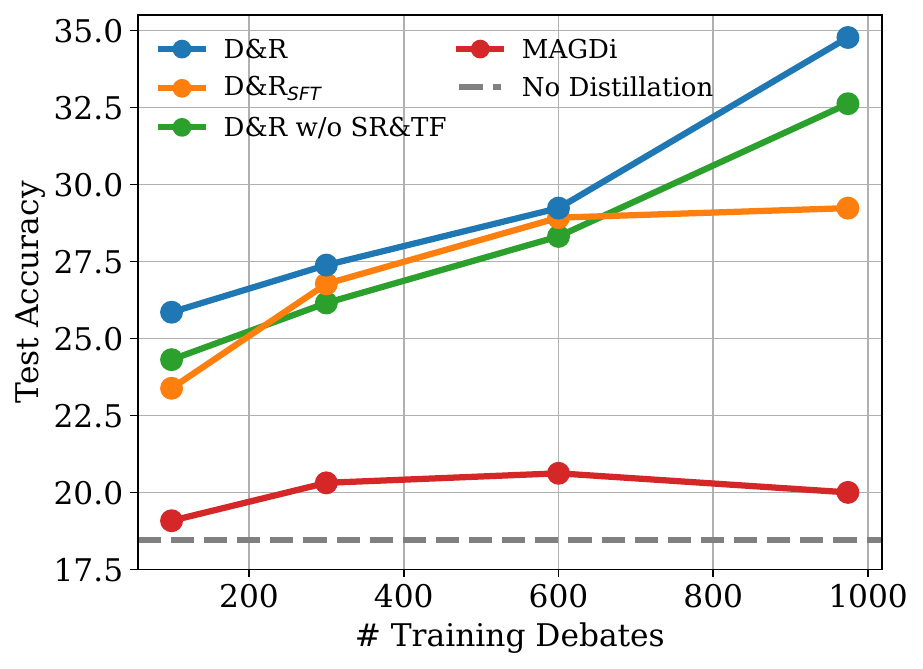}
    \vspace{-0.2cm}
    \caption{Performance with different training debates on  MMLU Pro physics. The dashed line refers to the original Mistral-7B-Instruct without any distillation.}
    \label{fig:data_scale}
    \vspace{-0.3cm}
\end{figure}

\subsubsection{\mymodel~Works on Various Data Scales}
To further investigate \mymodel's adaptability to different training data size, we train \mymodel~on the physics category of MMLU Pro using different debate scales: 100, 300, 600, and 974 debates (full set). The results, reported in Figure \ref{fig:data_scale}, reveal the following key findings.

At first, our \mymodel~achieves effective distillation learning across all dataset sizes, while \textsc{MAGDi} struggles with smaller datasets and fails to distill knowledge and reasoning capabilities effectively. Then \mymodel~outperforms \textsc{MAGDi} with all data scales. And T-DPO always enhances the SFT models, proving that self-reflection and teacher feedback in T-DPO are crucial. Without them, performance declines. The performance of \mymodel~also improves as the data scale increases. Therefore, \mymodel~has better utilization of debate data and still can capture preference with limited data. It can collect multiple SFT data instances and preference trees from a single debate MAG, improving data efficiency. The structured information in our preference trees provides rich knowledge, while self-reflection and feedback data can help the student model deeply understand complex reasoning preferences. Hence \mymodel~can be applied in scenarios with limited data resources, achieving effective improvements with only a small number of debates. When given access to more data, it scales effectively, leading to further performance gains.

\begin{table}
  \centering
  \small
  \begin{tabular}{m{10ex}|lc}
      \toprule
      Distillation Data & Student Models & MATH \\
      \midrule
      \multirow{6}{10ex}{The Same Training Debates for \mymodel} & Mistral-7B-Instruct & 8.02 \\
      & \hspace{2ex}with \mymodel$_{SFT}$ & 17.64 (+9.62) \\
      & \hspace{2ex}with \mymodel & 17.32 (+9.30) \\
      \cmidrule{2-3}
      & Llama-3.1-8B-Instruct & 45.78 \\
      & \hspace{2ex}with \mymodel$_{SFT}$ & 47.06 (+1.28) \\
      & \hspace{2ex}with \mymodel & 48.02 (+2.24) \\
      \bottomrule
  \end{tabular}
  \caption{\label{tbl:llama_math}
    Comparison between different student models. The performance of distillation can be affected by the capability of student models. Llama-3.1-8B-Instruct is able to learn more during \mymodel~distillation.
  }
  \vspace{-0.4cm}
\end{table}

\subsubsection{\mymodel~with Different Student Models} \label{exp_stu_model}
\mymodel~achieves improvements across student models of varying capabilities. The effectiveness of distillation depends not only on teacher models and algorithms but also on the capability of our student model. MATH is a challenging dataset that requires complex reasoning. And the basic performance of Mistral-7B-Instruct on MATH is a little weak. As shown in Table \ref{tbl:main_results}, while \mymodel$_{SFT}$ achieves a strong distillation performance on the MATH dataset, \mymodel~does not provide further improvements. To investigate whether this limitation stems from the student model’s capacity, we distill the same training debates into a more powerful student model, Llama-3.1-8B-Instruct. The results, presented in Table \ref{tbl:llama_math}, confirm this hypothesis. On the MATH dataset, where Mistral may reaches a distillation bottleneck, Llama-3.1-8B-Instruct continues to improve with both SFT and T-DPO in \mymodel. The complex reasoning capacity of Mistral-7B-Instruct might not be sufficient for full distillation. It can absorb teacher models’ reasoning capabilities during SFT but reaches its capacity limit, preventing further gains from T-DPO. On the contrary, Llama-3.1-8B-Instruct is stronger and has a much better basic MATH performance than Mistral-7B-Instruct. It has much potential for distillation. Therefore, during the T-DPO, it can still learn more complex reasoning capabilities from our training debates, demonstrating that \mymodel~can work well on the MATH dataset and fit different student models. That proves \mymodel's robust effectiveness across diverse student models.

\subsubsection{The Influence of Different Objectives}
Our proposed T-DPO objective is well-suited for distilling reasoning abilities from debates. In reasoning tasks, various preference-learning objectives exist \citep{xiong_building_2025, zhang_chain_2024}. In this section, we investigate whether T-DPO is sufficient and if other training objectives could replace T-DPO in \mymodel. RPO \citep{pang_iterative_2024} is a DPO-based training objective designed for reasoning tasks. Previous experiments \citep{pang_iterative_2024} suggest that the sequence-level log probabilities for chosen completions with RPO are increasing, while the standard DPO suffers a decrease.

\begin{table}[t]
  \centering
  \small
  \begin{tabular}{l*{2}{c}ccc}
      \toprule
      \multirow{2}{*}{Variants} & \multicolumn{3}{c}{MMLU Pro} & \multirow{2}{*}{MATH} & \multirow{2}{*}{Avg.} \\
      & CS & Phys. & Biol. & \\
      \midrule
      \textit{w/ SFT} & 32.68 & 29.23 & 64.44 & 17.64 & 36.00 \\
      \textit{w/ RPO} & \textbf{36.59} & 32.00 & 66.95 & 17.00 & 38.14 \\
      \textit{w/ T-DPO} & 33.17 & \textbf{34.77} & \textbf{67.36} & \textbf{17.32} & \textbf{38.16} \\
      \bottomrule
  \end{tabular}
  \vspace{-0.2cm}
  \caption{\label{tbl:diff_obje}
Comparison among \mymodel ~variants. Although \textit{w/ RPO} is better than \textit{w/ SFT}, it generally performs worse than \textit{w/ T-DPO} version.
  }
   \vspace{-0.2cm}
\end{table}
To compare these objectives, we apply RPO to our preference trees and report the results in Table \ref{tbl:diff_obje}. Following \citet{pang_iterative_2024}, we set the hyperparameter $\alpha$ in the RPO loss as 1. We observe that \mymodel$_{RPO}$ performs better than \mymodel$_{SFT}$ and close to \mymodel$_{T-DPO}$ in average performance. However, according to our experimental logs, the log probability does not show a consistent trend across different data categories. In some categories, such as physics, RPO increases log probabilities beyond T-DPO, yet performance does not improve. In others like computer science, log probabilities show little difference from T-DPO, yet RPO performs well. There is no clear correlation between log probability trends and model performance, suggesting that RPO’s effectiveness may be task-dependent. In conclusion, based on preference trees from debate graphs, T-DPO remains the most suitable objective for debate distillation.

\section{Conclusion}
In this paper, we proposed \mymodel, a novel distillation framework to enhance small models through debates. 
The debate process, involving teacher and student models, fully exposed the student model’s weaknesses, providing self-reflection and feedback. 
Then, we used T-DPO with preference trees from debate MAGs, capturing structured information and preference relationships between different reasoning paths. \mymodel~effectively distilled the teacher models' knowledge and reasoning capabilities into the student model through SFT and T-DPO. Our experiments demonstrate that \mymodel~achieves superior distillation compared to baseline methods, bridging the gap between smaller student models and larger teacher models. Compared to online debate everytime, \mymodel~significantly improved inference efficiency. Beyond reasoning, it also strengthened the student model's ability to self-correct. In future, we will explore long-term learning where the student model continues to improve over time through continuous debates and iterative feedback. 

\section*{Acknowledgements}
This research was supported by the National Research Foundation, Singapore under its
National Large Language Models Funding Initiative (AISG Award No: AISG-NMLP-2024-002).
Any opinions, findings and conclusions or recommendations expressed in this material are those
of the author(s) and do not reflect the views of National Research Foundation, Singapore. It was also supported by the National Natural Science Foundation of China (No. U21B2009) and the Lee Kong Chian Fellowship. This work was done during a visit to Singapore Management University funded by the China Scholarship Council.

\section*{Limitations}

% Since December 2023, a "Limitations" section has been required for all papers submitted to ACL Rolling Review (ARR).
Our study has limitations in two aspects. First, while multi-agent debate and preference learning are explored across a wide range of tasks, such as evaluation and model alignment, our work focuses solely on the knowledge and reasoning tasks. Investigating \mymodel~in other tasks would be valuable for assessing its generalizability. Second, our evaluation primarily measures the correctness of final answers rather than the validity of reasoning paths, which may lead to some false positive results. Process verification remains an open challenge, and we anticipate future studies that incorporate comprehensive evaluations of both reasoning paths and final outcomes.

\bibliography{custom, research}

\clearpage
\appendix

\begin{table*}
  \centering
  \small
  \begin{tabular}{lcccccccc}
      \toprule
      Datasets & \makecell{\# Training\\Debates} & \makecell{Round\\(1/2/3/4)} & \# R & \makecell{\# SR} & \# TF & \makecell{\# SFT} & \makecell{\# T-DPO} & \makecell{\# Test\\Problems} \\
      \midrule
      MATH & 1000 & 85/606/149/160 & 9524 & 2579 & 4111 & 6957 & 7744 & 5000 \\
      Physics & 974 & 136/696/104/38 & 7968 & 1722 & 3016 & 6246 & 4238 & 325 \\
      Biology & 478 & 151/295/24/8 & 3380 & 553 & 1061 & 2827 & 1214 & 239 \\
      Computer Science & 205 & 32/151/17/5 & 1620 & 340 & 554 & 1280 & 670 & 205 \\
      \bottomrule
  \end{tabular}
  \vspace{-0.2cm}
  \caption{\label{tbl:data_stat}
    Detailed statistics of our datasets. Round (1,2,3,4) shows the number of debates with 1/2/3/4 rounds. R, SR, and TF represent Responses, Self-Reflection, and Teacher Feedback, respectively. \# SFT is the number of SFT data constructed from debates. \# T-DPO is the number of preference trees constructed from debates for T-DPO.
  }
  \vspace{-3ex}
\end{table*}

\begin{table}
  \centering
  \small
  \begingroup
  \setlength{\tabcolsep}{4pt}
  \begin{tabular}{lcccc}
      \toprule
      \multirow{2}{*}{Models} & \multicolumn{3}{c}{MMLU Pro} & \multirow{2}{*}{Avg.} \\
      & CS & Physics & Biology & \\
      \midrule
      Mistral & 601.87 & 692.51 & 586.92 & 627.10 \\
      \textsc{MAGDi} & 578.05 & 619.64 & 533.54 & 577.08 \\
      ST$_{Claude}$ & 555.38 & 550.27 & 556.16 & 553.94 \\
      \mymodel & 520.24 & 584.17 & 482.07 & 528.83 \\
      Debate & 17993.04 & 21789.26 & 10526.94 & 16769.75 \\
      \bottomrule
  \end{tabular}
  \endgroup
  \caption{\label{tbl:token_cost}
    Token cost on the three categories of MMLU Pro. For each category, we report the averaged per-problem token cost on test sets.
  }
  \vspace{-0.2cm}
\end{table}

\begin{table}
    \centering
    \small
    \begin{tabular}{lccc}
        \toprule
        \multirow{2}{*}{Models} & \multicolumn{3}{c}{MMLU Pro} \\
         & CS & Physics & Biology \\
        \midrule
        GPT-4o & 75.12 & 74.77 & 88.70 \\
        Claude 3.5 & 77.07 & 77.54 & 87.87 \\
        Gemini 1.5 Pro & 78.05 & 76.31 & 84.94 \\
        Debate & 84.29 & 85.23 & 90.38 \\
        Llama-3.1-8B-Instruct & 46.83 & 39.38 & 65.27 \\
        \hspace{2ex}with \mymodel$_{SFT}$ & 50.24 & 44.92 & 71.97 \\
        \hspace{2ex}with \mymodel & 52.68 & 46.46 & 74.06 \\
        \bottomrule
    \end{tabular}
    \caption{\label{tbl:more_results_mmlupro}
        Results of teacher models, debates, and a more powerful student model, Llama-3.1-8B-Instruct, on the three categories of MMLU Pro.
    }
    \vspace{-0.2cm}
\end{table}

\section{Experimental Setting Details} \label{app_experimental_setting_details}

\paragraph{Datasets}
The data scale varies significantly across different categories in MMLU Pro. To ensure a sufficient amount of test data, we adopt different training-to-test split ratios: 1:1 for computer science, 3:1 for physics, and 2:1 for biology. When structuring information in the root nodes of preference trees, we filter out failed responses where the content is missing.

Table \ref{tbl:data_stat} presents the detailed statistics of our dataset. Most of our debates can proceed to the second round, resulting in meaningful interactions between the agents. In contrast, prior debate datasets contain a large proportion of debates that conclude after the initial round without further interaction. For the statistics of the previous debate data, please refer to \citet{chen_magdi_2024}. With self-reflection and teacher feedback, agents in our framework can promptly correct errors after the second round, ensuring more consensual correct answers and reducing instances where debates terminate without reaching a valid solution. Furthermore, the number of SFT and T-DPO training instances generated from our debates is 2.5–7.7 times the number of debates, highlighting \mymodel’s efficient utilization of debate data for distillation.

\paragraph{Models}
To ensure the reproducibility of our work, we use the following versions of the teacher and student models: \texttt{gpt-4o-2024-08-06} for GPT-4o, \texttt{claude-3-5-sonnet-20241022} for Claude 3.5, \texttt{gemini-1.5-pro-002} for Gemini 1.5 Pro, and \texttt{Mistral-7B-Instruct-v0.2} for Mistral-7B-Instruct.

\paragraph{Implement Details}
For debate generation, we use the following generation configurations: \texttt{temperature} is 0.3, the hard limit \texttt{max tokens} is 700 (1000 for MATH and we also specify a soft limit in the prompt templates). For SFT, we train the student model on all datasets for 2 epochs with \texttt{learning rate} as 2e-4 and \texttt{batch size} as 16.

For T-DPO, we train the student model on MMLU Pro for 3 epochs with \texttt{learning rate} as 5e-6 and \texttt{batch size} as 16. On MATH, we train the student model for 1 epoch with \texttt{learning rate} as 1e-6 and \texttt{batch size} as 16. We limit the total sequence length to 2000 and the prompt sequence length to 1400. For LoRA, we set the LoRA adapter rank as 16 and the alpha as 32.

For evaluation, our generation configurations are \texttt{temperature} as 0.3, \texttt{top\_p} as 0.9, and the hard limit \texttt{max tokens} as 700 (1000 for MATH). For the reimplementation of \textsc{MAGDi}, we follow its original training hyperparameters and generation configurations.

\section{Quality Analysis of Debate Data}

\begin{figure*}
    \centering
    \includegraphics[width=0.89\linewidth]{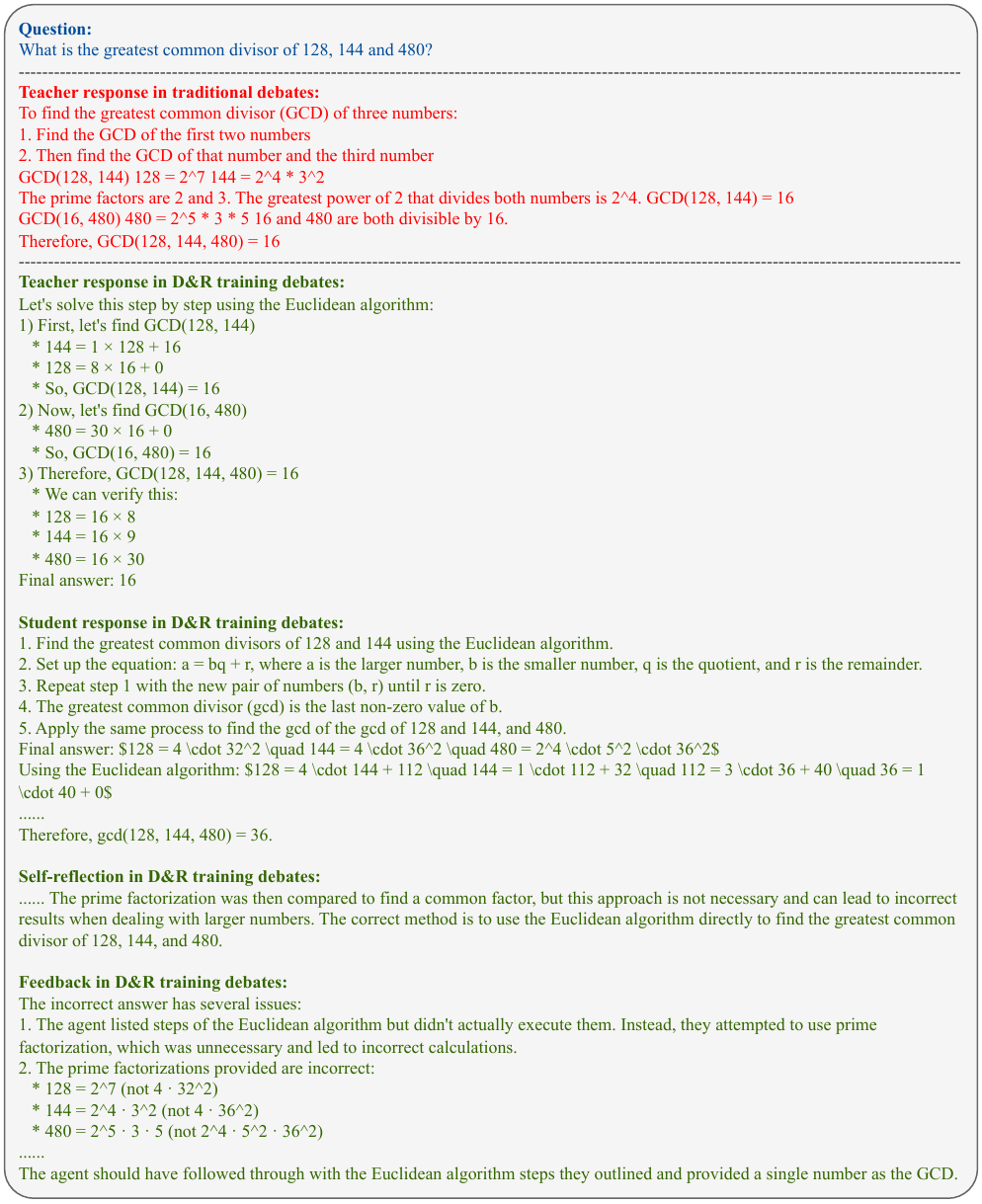}
    \caption{An example from the MATH dataset. Given the question, we show a response from \citet{chen_magdi_2024}, which is generated with a traditional debate method without any self-reflection, feedback, and student models, and then two responses, a self-reflection, and feedback from our \mymodel~training debates. Our debate data has a better quality with detailed responses, self-reflection, and feedback.}
    \label{fig:debate_case}
\end{figure*}

Our debate data exhibits higher quality than traditional debate data, providing clearer reasoning structures and more targeted feedback. In \mymodel, we introduce an improved debate data generation method that actively involves the student model and explicitly incorporates self-reflection and feedback from teacher models. As a result, our training debate data is enriched with detailed reasoning and targeted guidance. To illustrate the advantages of our approach, Figure \ref{fig:debate_case} presents a comparison between our debate data and traditional debate data from MATH.

For the problem of finding the greatest common divisor, the teacher model in our debate data clearly outlines the use of the Euclidean algorithm, systematically performs the calculations, and verifies the result. In contrast, the debate data from \citet{chen_magdi_2024}, constructed using a traditional approach, lacks an explicit mention of the prime factorization method and presents a less structured response.

In our debate, the student model initially produces an incorrect response. While it introduces the Euclidean algorithm, it actually applies prime factorization, making multiple calculation errors in the process. However, the inclusion of self-reflection and teacher feedback highlights these mistakes and provides explicit recommendations for correctly applying the Euclidean algorithm. This guidance plays a crucial role in shaping subsequent debate rounds. Traditional multi-agent debate frameworks provide agents only with prior responses, requiring them to independently compare answers and identify correct and incorrect elements. This process adds cognitive complexity, making it harder for the agents to reason effectively. In contrast, our method streamlines error correction by integrating self-reflection and feedback, offering clear suggestions for improvement. Consequently, in the following debate rounds, agents can more efficiently rectify previous mistakes and generate more accurate responses, leading to a more effective learning process.

\begin{figure}
    \centering
    \includegraphics[width=0.88\columnwidth]{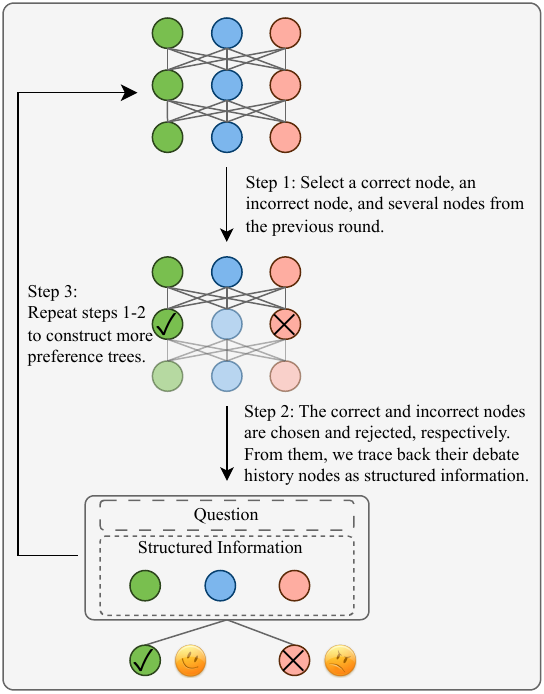}
    \caption{The schematic diagram showing how to construct preference trees from debates.}
    \label{fig:schematic_diagram}
    % \vspace{-0.5cm}
\end{figure}

\section{Token Cost} \label{app_token_cost}
\vspace{-0.2cm}
We report averaged per-problem token cost on the test sets of three categories of MMLU Pro in Table \ref{tbl:token_cost}, which is used to calculate efficiency in Section \ref{exp_efficiency}. \mymodel~learns to reach answers with better reasoning paths and reduce failures during generation, hence has a better efficiency than Mistral, \textsc{MAGDi}, and ST$_{Claude}$. Since knowledge and reasoning capabilities have been distilled into the student model, there is no need to use multiple large models for debate. The token cost of \mymodel~is much lower than that of debate.

\section{More Results on MMLU Pro}
% \vspace{-1ex}
In Table \ref{tbl:more_results_mmlupro}, we report more results on the three categories of MMLU Pro, including three teacher models, debates, and Llama-3.1-8B-Instrct. The teacher models and debates achieve strong performance, ensuring the quality of distillation data. Besides MATH (as shown in Section \ref{exp_stu_model}), Llama-3.1-8B-Instruct also gain improvements on MMLU Pro. It demonstrates the broad applicability and robustness of \mymodel.

\section{The Detailed Schematic Diagram for Constructing Preference Trees}
% \vspace{-1ex}
In Figure \ref{fig:schematic_diagram}, we present a schematic diagram to complement Figure \ref{fig:overview}, illustrating the steps to construct preference trees:
\begin{enumerate}
    \item Select a correct node, an incorrect node, and several nodes from the previous round.
    \vspace{-0.1cm}
    \item The correct and incorrect nodes are chosen and rejected, respectively. From them, we trace back their debate history nodes as structured information.
    \vspace{-0.1cm}
    \item Repeat steps 1-2 to construct more preference trees.
\end{enumerate}

\section{Designed Prompt Templates in \mymodel} \label{app_temp}
\begin{figure*}
    \centering
    \includegraphics[width=0.89\linewidth]{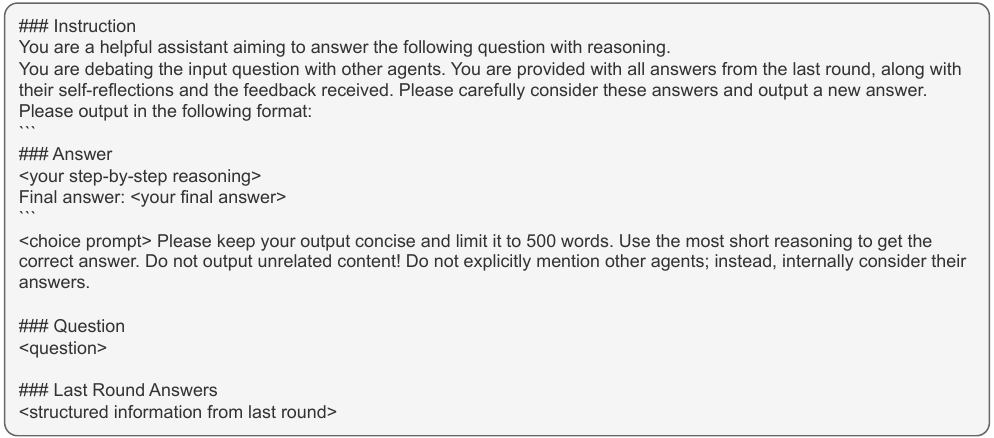}
    \caption{The prompt template for generating debates.}
    \label{fig:debate_temp}
\end{figure*}

\begin{figure*}
    \centering
    \includegraphics[width=0.89\linewidth]{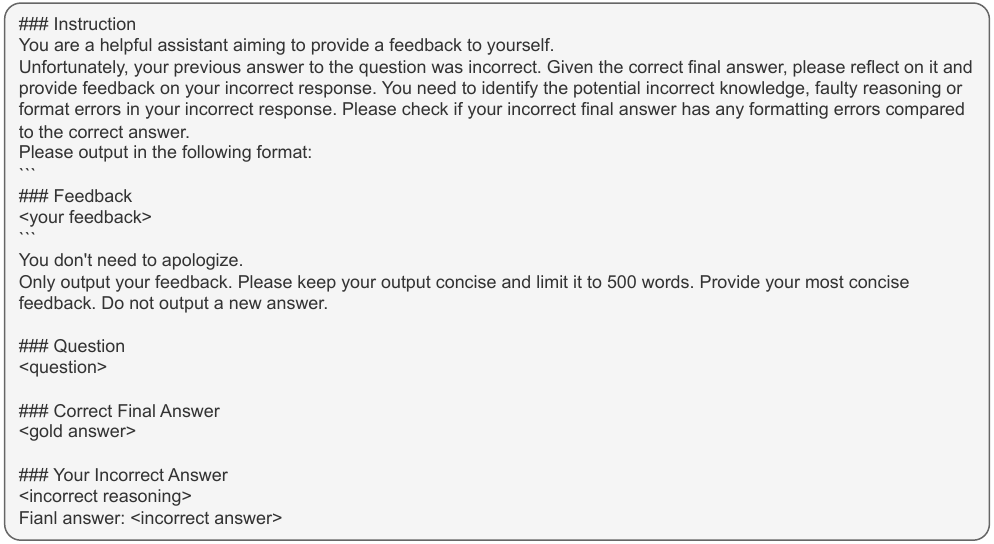}
    \caption{The prompt template for self-reflecting in debates.}
    \label{fig:reflect_temp}
\end{figure*}

\begin{figure*}
    \centering
    \includegraphics[width=0.89\linewidth]{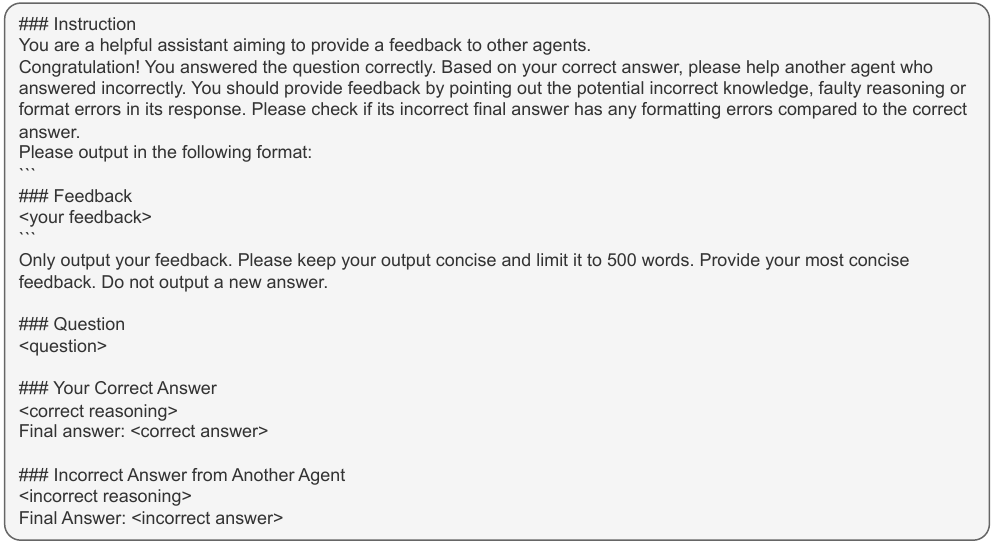}
    \caption{The prompt template for generating feedback in debates.}
    \label{fig:feedback_temp}
\end{figure*}

\begin{figure*}
    \centering
    \includegraphics[width=0.89\linewidth]{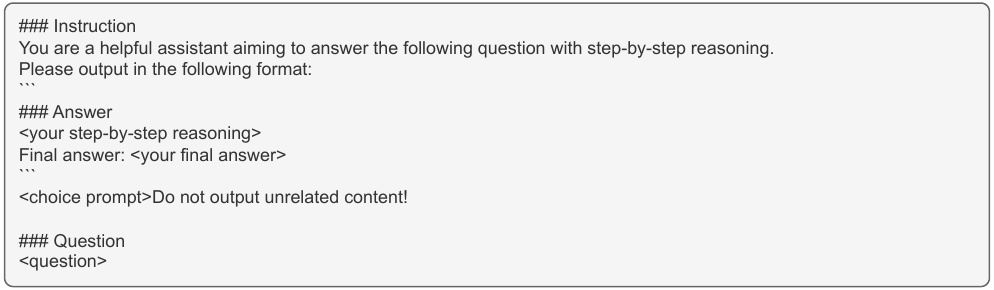}
    \caption{The prompt template for \mymodel~distillation.}
    \label{fig:distill_temp}
\end{figure*}

In this section, we provide the prompt templates that are used in our \mymodel~for generating debates and distillation in Figure \ref{fig:debate_temp}, \ref{fig:reflect_temp}, \ref{fig:feedback_temp}, and \ref{fig:distill_temp}.


\end{document}